\documentclass[10pt,twocolumn,letterpaper]{article}

\usepackage{iccv}              

\definecolor{iccvblue}{rgb}{0.21,0.49,0.74}
\usepackage[pagebackref,breaklinks,colorlinks,allcolors=iccvblue]{hyperref}

\def\paperID{171} 
\def\confName{ICCV}
\def\confYear{2025}

\title{PointDiffuse: A Dual-Conditional Diffusion Model for Enhanced Point Cloud Semantic Segmentation}

\author{
  \textbf{Yong He}$^{1}$, 
  \textbf{Hongshan Yu}$^{2*}$, 
  \textbf{Mingtao Feng}$^3$, 
  \textbf{Tongjia Chen}$^4$,
  \textbf{Zechuan Li}$^{2}$,\\
  \textbf{Anwaar Ulhaq}$^5$, 
  \textbf{Saeed Anwar}$^6$, 
  \textbf{Ajmal Saeed Mian}$^4$ \\
  $^1$Anhui University, 
  $^2$Hunan University, 
  $^3$Xidian University,\\
  $^4$University of Western Australia,
  $^5$Central Queensland University,
  $^6$Australian National University,
  \\
}

\begin{document}
\maketitle
\begin{abstract}
Diffusion probabilistic models are traditionally used to generate colors at fixed pixel positions in 2D images. Building on this, we extend diffusion models to point cloud semantic segmentation, where point positions also remain fixed, and the diffusion model generates point labels instead of colors. To accelerate the denoising process in reverse diffusion, we introduce a noisy label embedding mechanism. This approach integrates semantic information into the noisy label, providing an initial semantic reference that improves the reverse diffusion efficiency. Additionally, we propose a point frequency transformer that enhances the adjustment of high-level context in point clouds. To reduce computational complexity, we introduce the position condition into MLP and propose denoising PointNet to process the high-resolution point cloud without sacrificing geometric details. Finally, we integrate the proposed noisy label embedding, point frequency transformer and denoising PointNet in our proposed dual conditional diffusion model-based network (PointDiffuse) to perform large-scale point cloud semantic segmentation. Extensive experiments on five benchmarks demonstrate the superiority of PointDiffuse, achieving the state-of-the-art mIoU of 74.2\% on S3DIS Area 5, 81.2\% on S3DIS 6-fold and 64.8\% on SWAN dataset. 
\end{abstract}    
\vspace{-2mm}
\section{Introduction}
\label{sec:intro}

Point cloud segmentation is a fundamental 3D vision task supporting a wide range of applications in autonomous driving~\cite{yuan2024ad,qian2024nuscenes}, and robotics~\cite{christen2023learning,stathoulopoulos2023frame}. Semantic segmentation methods have evolved from early MLP-based methods~\cite{qian2022pointnext,deng2023pointvector,lin2023meta} that apply MLPs on points to model local point features, to convolution-based methods~\cite{he2023danet,thomas2024kpconvx} that learn convolutional weights to model local point connections. More recently, transformer-based methods~\cite{wu2022point,he2023full} have been proposed to incorporate the attention mechanism to learn contextual information.

However, raw point clouds captured by depth sensors or LiDARs are often contaminated with noise due to environmental factors or sensor limitations. 
While local aggregation methods exhibit strong representational capabilities, they remain vulnerable to noise, particularly at object boundaries and in small object regions. Consequently, the denoising limitations of existing segmentation methods can lead to suboptimal performance in these challenging areas. 
Recently, the Diffusion Probabilistic Model (DPM)~\cite{ho2020denoising} has gained popularity as a powerful class of generative models capable of generating
high-quality and diverse images. DPMs have an inherent denoising capability. Motivated by its success, many researchers have applied DPMs in several other 2D computer vision tasks, such as image segmentation~\cite{xu2023open,wu2023diffumask}, and object detection~\cite{chen2023diffusiondet,bansal2023universal}. Diffusion models have also been gradually extended to 3D computer vision tasks such as shape generation~\cite{mo2024dit,zheng2023locally}, shape completion~\cite{zhou20213d,cheng2023sdfusion}, 3D detection~\cite{hou20243d}, and 3D part segmentation~\cite{wu2023sketch}. 
Diffusion models are particularly effective at handling noise, given their strong denoising capabilities. Compared to traditional local aggregation networks, diffusion models offer two key advantages: 
\begin{itemize}
    \item \emph{Enhanced flexibility}: Diffusion models exhibit enhanced flexibility and can generate diverse predictions through multiple runs. This is akin to test-time augmentation, as demonstrated in~\cite{lai2022stratified}, where diverse predictions are combined to produce better segmentation.
    \item \emph{Improved adaptability}: Multiple runs of generative models enable them to adapt to different scenarios or data distributions, improving their generalization ability.
\end{itemize}

Given the above advantages, we argue that diffusion models are particularly well-suited for handling noisy point cloud data characterized by intricate contextual connections.  Motivated by this,  we endeavor to leverage diffusion for large-scale point cloud segmentation. However, using diffusion models for this task is not straightforward and faces the following challenges. 
\begin{itemize}
    \item \emph{Long runtime}: Generating large-scale point labels requires a large number of denoising steps in the reverse diffusion process, leading to \textcolor{black}{extended inference time.}
    \item \emph{Limited local denoising ability}: The ability of diffusion models to handle local noise is somewhat restricted when processing unordered and irregular points.
     \item \textcolor{black}{\emph{Loss of fine-grained features}: The diffusion model may overly smooth the label distribution, leading to the loss of subtle semantic features.}
\end{itemize}
\vspace{0.5mm}

To overcome the aforementioned challenges, we propose a dual conditional diffusion network for point cloud segmentation, coined \textbf{PointDiffuse}. Our key idea involves integrating dual conditions (i.e., semantic and position conditions) and local aggregation into the diffusion model.  
Specifically, \textbf{i)} we design a noisy embedding layer for low-level geometric denoising, which performs convolution operations on noisy features. The convolutional weights are learned from the semantic condition and used to mask the noise based on the position condition. By anchoring the noise using dual conditions, the diffusion model obtains a preliminary semantic reference, facilitating the rapid generation of point labels while suppressing the loss of \textcolor{black}{fine-grained features}. \textbf{ii)} We propose a point frequency transformer for high-level context denoising. The point frequency transformer converts the noise features and position conditions from the spatial to the frequency domain and then uses vector attention on these features to learn context connections. The diffusion model achieves position awareness by introducing the position condition in the transformer \textcolor{black}{to improve the local denoising ability.} \textbf{iii)} Furthermore, to decrease the long runtime, we propose a simple but efficient Denoising PointNet over the high resolution point clouds that introduces the position condition into MLP.
In summary, our contributions are:
\vspace{0.5mm}
\begin{itemize}
\item We propose \emph{PointDiffuse}, a dual conditional diffusion model 
for large-scale point cloud segmentation.
\item We propose a \emph{Noisy Label Embedding} mechanism that integrates {semantic and local position conditions} to stabilize variance in the diffusion process.
\item We propose a \emph{Point Frequency Transformer} that adaptively filters out noise, enhancing segmentation quality. 
\item We develop a simple yet effective \emph{Denoising PointNet} to improve the overall computational efficiency.
\end{itemize}
\vspace{0.5mm}

To demonstrate the effectiveness of our proposed method, we conduct extensive experiments on five benchmark datasets: S3DIS~\cite{armeni20163d}, ScanNet~\cite{dai2017scannet}, SWAN~\cite{ibrahim2023sat3d}, SemanticKITTI~\cite{behley2019semantickitti}, and ShapeNet~\cite{yi2016scalable}. Without relying on multi-dataset or multi-modal joint training, our method achieves state-of-the-art performance with 74.2\% and 81.2\% mIoU on S3DIS Area 5 and 6-fold cross-validation for indoor semantic segmentation, and 64.8\% mIoU on SWAN for outdoor semantic segmentation. It also achieves competitive performance of 88.4\% Ins. mIoU on ShapeNet for object part segmentation.

\section{Related Work}
\label{relatedwork}

\vspace{-1mm}
\textbf{{MLP-Based Approaches}}: PointNet~\cite{qi2017pointnet} is a milestone in deep learning for point clouds. It employs shared MLPs to leverage point-wise features and a symmetric function such as max-pooling to aggregate these features into global representations. However, the network struggles to exploit local features. To address this issue, hierarchical architectures have been proposed to aggregate local features with MLPs~\cite{qi2017pointnet++,li2018so} such that the model can benefit from efficient sampling and grouping of the point set. Recent works have focused on enhancing point-wise features by hand-crafting geometric connections~\cite{ma2021rethinking,ran2022surface,deng2023pointvector} or graphs~\cite{zhao2019pointweb,feng2020point,xu2021learning}. {These MLP-based approaches are simple and effective with fewer learnable parameters. 

\noindent\textbf{{Convolution-Based Approaches}}: To improve local feature learning, various point convolutions~\cite{wang2018deep,wu2019pointconv,wang2019dynamic}, inspired by 2D grid convolution, dynamically learn convolutional weights (i.e. dynamic kernels) through weight functions from local point geometric connections. KCNet~\cite{shen2018mining}, KPConv~\cite{thomas2019kpconv} and KPConvX~\cite{thomas2024kpconvx} predefine a set of fixed kernels (coined kernel points) in the local receptive field and then learn the weights on these kernels from the geometric connections between local points and kernel points using Gaussian and linear correlation functions, respectively. Another technique~\cite{wu2019pointconv,Lei_2020_CVPR} associates coefficients with kernels to further adjust the learned weights, where the coefficients are also derived from point coordinates. Point convolution learns the convolution weights from the low-level point coordinates, lacking high-level semantic awareness. 
Point convolution methods excel at capturing local features in the initial layers but fail to capture the global context.

\noindent\textbf{{Transformer-Based Approaches}}: To improve semantic awareness and global context learning, point transformers learn attention weights from the feature relationships (i.e., the connections between the query and key) and position differences (i.e., point position encoding). For example, Point Transformer~\cite{zhao2021point} introduces the local vector attention function, to the local points, which learns the weights from the feature differences masked by position encoding. Point Transformer V2~\cite{wu2022point}, makes the position encoding more complex by introducing a position encoding multiplier to adjust the features. Similarly, the Full Point Transformer~\cite{he2023full} operates the position encoding over the global and local receptive fields alternatively, enabling more complex positional encoding. 
Compared to MLPs and point convolutions, point transformers capture semantic information more effectively. Although local point aggregation methods like MLP, Point Convolution, and Transformer exhibit strong learning capabilities in clean point clouds, they encounter difficulty extracting robust features from noisy point clouds due to their lack of explicit denoising ability.

 
\noindent \textbf{{Diffusion Based Approaches:}} Diffusion models~\cite{ho2020denoising,nichol2021improved} belong to a class of generative models grounded in Markov chains that gradually recover data samples through a denoising process. These models have been successful in many generative tasks, spanning image~\cite{rombach2022high,epstein2023diffusion} and point cloud generation~\cite{tyszkiewicz2023gecco, mikuni2023fast}. Beyond their achievements in generative tasks, diffusion models hold substantial promise for perception tasks such as natural image segmentation~\cite{liu2023diffusion,xu2023open}, medical image segmentation~\cite{wu2024medsegdiff,wu2024medsegdiffv2} and point part segmentation~\cite{wu2023sketch}.
\textcolor{black}{Recent works \cite{zheng2024point, liu20243d} generate points from noisy point clouds to help the backbone capture geometric priors for semantic segmentation. In contrast, our method directly generates {\em point labels} from noise for semantic segmentation.} 
In this paper, we integrate local point aggregation techniques into the diffusion model and introduce several novel local aggregation modules with denoising capabilities. Building upon our novel local point aggregation techniques, we develop a diffusion model-based semantic segmentation network for large-scale point clouds.

\section{Method}
\label{method}



\label{subsec1_diffusion_process}
We consider a labeled point cloud $\{I_i, x_i^0\}\in\mathbb{R}^{N \times (3+M)}$, where $I_i\in\mathbb{R}^{N \times 3}$ is the point position,  $x_i^0\in\mathbb{R}^{N \times M}$ is the point label, expressed in one-hot form. Here, $N$ is the number of points and $M$ is the number of classes. The corresponding features for point $I_i$ are denoted as $F_i$. Similar to other diffusion models such as \cite{ho2020denoising}, PointDiffuse consists of two stages: a forward diffusion and a reverse diffusion. In the forward diffusion stage, Gaussian noise $\epsilon$ is gradually added to the point label $x_i^0$  as,
\vspace{-2mm}
\begin{equation}
\begin{aligned}
\label{eq:dm_forward}
q(x_i^{1:T}|x_i^0) &= \prod_{t=1}^T q(x_i^t |x_i^{t-1}),\\
q(x_i^t | x_i^{t-1}) &= \mathcal{N}(x_i^t; \sqrt{1-\beta_t}x_i^{t-1}, \beta_t I_i).
\end{aligned}
\end{equation}

\textcolor{black}{The forward process starts with the original point labels $x_i^0$ and progressively adds noise to it in $T$ steps, generating noisy point labels $x_i^T$. Therefore, $x_i^{t-1}$ and $x_i^t$ are the noisy point labels at step $t$ and $t-1$, respectively, in the forward process.} $q(x_i^t | x_i^{t-1})$ is a Gaussian transition kernel, which adds noise to the point label input $x_i^0$ with a variance schedule $\beta_1, ..., \beta_T$.

During reverse diffusion, a denoising model $\boldsymbol{\theta}$ is learned to recover the original point labels from noisy point labels 
by reversing the noise addition process. This reverse process can be represented as,
\vspace{-1mm}
\begin{equation}
\begin{aligned}
\label{eq:dm_reverse}
&p_{\boldsymbol{\theta}}(x_i^{0:T}|x_i^0) = p(x_i^T)\prod_{t=1}^T p_{\boldsymbol{\theta}}(x_i^{t-1} |x_i^{t}),\\
&p_{\boldsymbol{\theta}}(x_i^{t-1} | x_i^{t}; S_i, P_{ij}) = \mathcal{N}(x_i^{t-1}; \mu_{\boldsymbol{\theta}}, \rho_t^2 I_i).
\end{aligned}
\end{equation}

The reverse process is also Markovian with Gaussian transition kernels, which use fixed variances $\rho_t^2$. Here, $S_i$ represents the additional semantic condition, while $P_{ij}$ denotes the supplementary local position condition, with $j$ indicating the neighborhood point index of point $I_i$. \textcolor{black}{The reverse process starts from noisy point labels $x_i^T$ and gradually denoises them to recover the original point labels $x_i^0$ . Therefore, $x_i^t$ and $x_i^{t-1}$ represent the generated point labels at step $t$ and $t-1$, respectively, in the
reverse process.} The diffusion model is trained to reverse the forward process, ultimately generating the point label.
\vspace{-2mm}
\begin{equation}
\begin{aligned}
&x_i^{t-1} = \frac{1}{\sqrt{\alpha}_t}\big(x_i^t -  \frac{\beta_t}{\sqrt{1-\overline{\alpha}_t}} \epsilon_{\boldsymbol{\theta}}(x_i^t, S_i, P_{ij}, t)\big) + \sqrt{\beta_t}\epsilon,
\end{aligned}
\end{equation}
\noindent where $\alpha_t = 1- \beta_t$, and $\overline{\alpha}_t = \prod_{s=0}^T \alpha_s$, while $\epsilon_{\boldsymbol{\theta}}$ is the noise prediction network, implemented by a learnable network with the parameter set $\boldsymbol{\theta}$. Our diffusion model learns to match generated label $x_i^{t-1}$ with ground truth label $x_i^0$ by minimizing the difference between ground truth Gaussian noise $\epsilon$ and predicted noise $\epsilon_{\boldsymbol{\theta}}(x_i^t, S_i, P_{ij},t)$.

\begin{figure*}[t]
\centering
\includegraphics[width= 0.95 \textwidth]{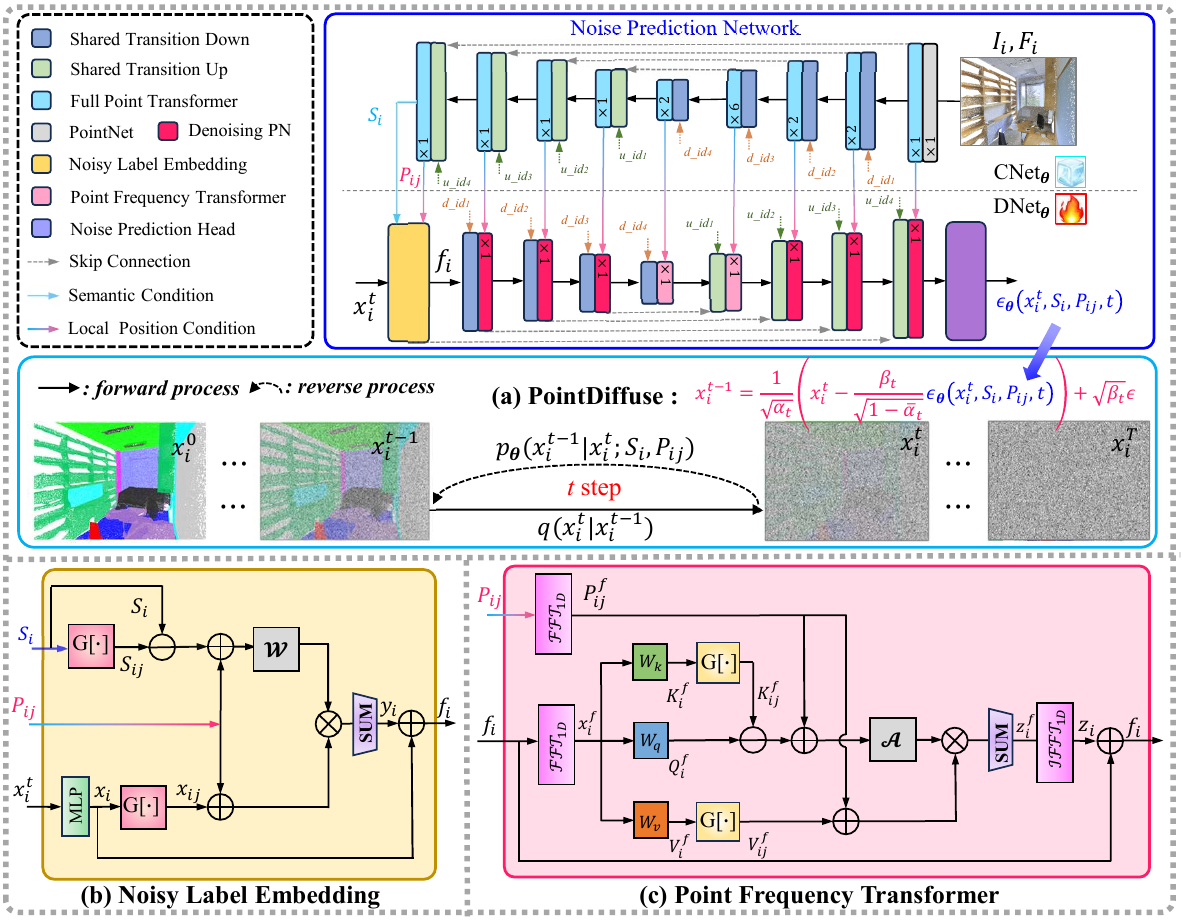}
\caption{An illustration of PointDiffuse, starting with (a) an overview of the entire pipeline, followed by detailed views of its key sub-modules: (b) Noisy Label Embedding for variance stabilization and (c) Point Frequency Transformer for noise filtering.}
\vspace{-2mm}
\label{fig:archi}
\end{figure*}

\subsection{Overall Architecture}
\label{subsec2_architecture}
 
To further clarify the PointDiffuse network, we first introduce the single step $t$ of the diffusion process where we define the learnable network $\epsilon_{\boldsymbol{\theta}}$ over the input $x_i^t$ and conditions (i.e., semantic and position condition). The learnable \textit{Denoising Network} is expressed as,
\vspace{-2mm}
\begin{equation}
\label{eq:dm_denoising_network}
\epsilon_{\boldsymbol{\theta}}(x_i^t, S_i, P_{ij}, t) = {\rm DNet}_{\boldsymbol{\theta}}(x_i^t, S_i, P_{ij}, t).
\vspace{-2mm}
\end{equation}

\noindent 

Denoising Network $\rm DNet_{\boldsymbol{\theta}}$ is conditioned by the prior information, including semantic information $S_i$ and local position information $P_{ij}$. The prior information is learned by the \textit{Conditional Network}, which can be expressed as,
\vspace{-2mm}
\begin{equation}
\label{eq:dm_conditional_network}
S_i, P_{ij} = {\rm CNet}_{\boldsymbol{\theta}}(I_i, F_i).
\vspace{-2mm}
\end{equation}

Figure~\ref{fig:archi} shows the overall architecture of PointDiffuse. (a) The noise prediction network in PointDiffuse contains two sub-networks: a conditional network and a denoising network. The conditional network ${\rm CNet}_{\boldsymbol{\theta}}$ provides the semantic and local position conditions for the denoising network ${\rm DNet}_{\boldsymbol{\theta}}$, \textcolor{black}{where the semantic condition refers to the semantic information learned by the conditional network, pre-trained for semantic segmentation.}
In practice, the conditional network employs a U-Net structure. In our experiments, we employ the state-of-the-art Full Point Transformer network \cite{he2023full} for this. 
Similarly, the denoising network also adopts a U-Net structure, comprising one noisy label embedding layer, four encoding layers, four decoding layers, and one noise prediction head. Each of the first three encoding layers contains a Shared Transition Down and a Denoising PointNet, while the last encoding layer contains a Shared Transition Down and a Point Frequency Transformer. Each of the last three decoding layers contains a Shared Transition Up and a Denoising PointNet, while the first decoding layer contains a Shared Transition Up and a Point Frequency Transformer.
The Shared Transition Down/Up donates that Transition Down/Up blocks~\cite{zhao2021point} with the same encoding/decoding layer share identical neighborhood point indexes across the same resolution points. 

During training, we freeze the parameters of the pre-trained conditional network and train the denoising network. This training, at each step in the reverse process, uses a standard noise prediction loss $\mathcal{L}_n^t$ along with a label generation cross-entropy loss $\mathcal{L}{ce}$. The total loss becomes: 
\begin{equation}\label{eq:loss}
\begin{aligned}
\vspace{-2mm}
&\mathcal{L}_{total} = \gamma \mathcal{L}_n^t + (1-\gamma) \mathcal{L}_{ce}\\
&= \gamma||\epsilon-\epsilon_{\boldsymbol{\theta}}(x_i^t, S_i, P_{ij}, t)||^2 + (1-\gamma)||x_i^0 - x_i^{t-1}||^2,
\end{aligned}
\end{equation}

\noindent where $\epsilon \sim \mathcal{N}(0,I_i)$ and $\gamma$$\in$ (0,1) is a hyperparameter that balances the contribution of the loss terms.

\subsection{Noisy Label Embedding}%
High variance at each denoising step can lead to instability, requiring more diffusion steps to achieve the desired outcome, which ultimately results in suboptimal model performance. To overcome this, we propose a novel noisy label embedding module that leverages semantic and position information to condition the noisy label embedding, stabilizing the variance during diffusion. As shown in Figure~\ref{fig:archi} (b), the noisy label embedding on each point can be expressed as,
\vspace{-1mm}
\begin{equation}
\begin{aligned}\label{eq:noise_embedding}
&~~~~~~~~~~~~~~~~~~x_i = {\rm MLP}(x_i^t),\\
\vspace{-2mm}
y_i&=\sum_{j=1}^{K}\mathcal{W}\big((G[S_{i}] \ominus S_i ) \oplus P_{ij} \big) \otimes \big(G[x_{i}] \oplus P_{ij} \big), \\
&=\sum_{j=1}^{K}\mathcal{W}\big((S_{ij} \ominus S_i ) \oplus P_{ij} \big) \otimes \big(x_{ij} \oplus P_{ij} \big), \\
\vspace{-2mm}
&~~~~~~~~~~~~~~~~~~f_{i} = y_i \oplus x_{i},
\vspace{-2mm}
\end{aligned}
\end{equation}
where the summation aggregates the features over $K$ neighborhood points. $G[\cdot]$ is the grouping operation to group the global element (i.e., $S_i$ or $x_i$) into the local receptive field, yielding the local elements (i.e., $S_{ij}$ or $x_{ij}$). $\ominus, \otimes, \oplus$ are the element-wise subtraction, multiplication, and addition operations, respectively.

During noisy label embedding, we first use an $\rm MLP$ to project the noisy labels $x_i^t$ onto noisy  features $x_i$. 
Next, we integrate the semantic $S_i$ and local position condition $P_{ij}$ into noisy features to obtain enhanced noisy features $y_i$. 
In particular, we construct local semantic connections, incorporate local position information, and then learn the weights using a weight function $\mathcal{W}(\cdot)$ from these connections to adjust the noisy features. 
Finally, \textcolor{black}{we add the enhanced noisy feature \textcolor{black}{$y_i$} to the original noisy features \textcolor{black}{$x_i$} to \textcolor{black}{obtain the nosiy label embedding $f_i$, retaining important information that could potentially be lost.}}

Compared to classic point convolution, the proposed noisy label embedding offers two key advantages: 
\textbf{i)} Noisy label embedding learns the weights from the semantic conditions rather than simple coordination differences. Semantic conditions, derived from the final layer of the condition network, provide a rough but stable semantic reference for noisy features, significantly reducing the diffusion variance. 
In addition, this allows the semantic features to be further refined during denoising. 
\textbf{ii)} Inspired by the position encoding in the point transformer, we equip the semantic connection with the local position condition to be \textcolor{black}{better aware of the point cloud structure and further restrain the geometry detail loss.} This not only reduces the number of network training parameters but also ensures that both conditional and denoising networks perceive the same point positions. Such an alignment also helps prevent instability and performance degradation from inconsistent position encoding.

\subsection{Point Frequency Transformer}
The denoising network predicts the noise component from the noisy labels. While conventional point transformers effectively embed features, they struggle to learn noise in the spatial domain~\cite{ye2024diffusionedge}. \textcolor{black}{Motivated by this problem}, we propose a novel point frequency transformer that operates in the frequency domain given the noisy input feature $f_{i}\in\mathbb{R}^{N \times C}$ and the local position condition $P_{ij}\in\mathbb{R}^{N \times K \times C}$. 

Figure~\ref{fig:archi}~(c) shows our Point Frequency Transformer. First, we separately perform 1-D Fast Fourier Transform ($ \mathcal{FFT}_{\rm 1D}$) on the noisy features \textcolor{black}{$f_i$} along the $N$ dimension and on the local position condition \textcolor{black}{$P_{ij}$} along the $K$ dimension. The transformed noisy features $x_i^f$ and local position conditions $P_{ij}^f$ in the frequency domain are represented as,
\vspace{-2mm}
\begin{equation}
x_{i}^f = \mathcal{FFT}_{\rm 1D}(f_i), ~~~P_{ij}^f = \mathcal{FFT}_{\rm 1D}(P_{ij}).
\vspace{-2mm}
\end{equation}

We then learn the {query~$Q_{i}^f$, key~$K_{i}^f$, value~$V_{i}^f$} from these transformed noisy features,
\vspace{-2mm}
\begin{equation}
Q_{i}^f = W_q(x_{i}^f), ~K_{i}^f = W_k(x_{i}^f), ~V_{i}^f = W_v(x_{i}^f), 
\end{equation}
where $W_q, W_k, W_v$ are the project functions. The point frequency transformer is,
\vspace{-2mm}
\begin{equation}
\begin{aligned}
{z}_{i}^f&=\sum_{j=1}^{K}\mathcal{A} \big((G[K_{i}^f] \ominus Q_i^f) \oplus P_{ij}^f  \big) \otimes \big(G[V_{i}^f] \oplus P_{ij}^f \big), \\
&=\sum_{j=1}^{K}\mathcal{A}\big((K_{ij}^f \ominus Q_i^f) \oplus P_{ij}^f  \big) \otimes \big(V_{ij}^f \oplus P_{ij}^f \big), \\
\end{aligned}
\end{equation}
where $\mathcal{A}(\cdot)$ is the attention weight function, implemented by an MLP followed by a softmax. We project the feature learned by the transformer from the frequency domain 
back to the spatial domain by a 1-D Inverse Fast Fourier Transform ($ \mathcal{IFFT}_{\rm 1D}$). Finally, we use a residual connection from $f_{i}$ to avoid missing useful information.
\vspace{-2mm}
\begin{equation}
\begin{aligned}
{z}_{i} &= \mathcal{IFFT}_{\rm 1D}({z}_{i}^f),\\
f_{i} &= {z}_{i}\oplus f_i.
\end{aligned}
\end{equation}

\vspace{-2mm}
\textcolor{black}{By mapping features to the frequency domain, the point frequency transformer can separate low-frequency components, which represent the underlying structure, from high-frequency components, which are often associated with noise. This separation enables the model to more effectively distinguish between critical features and noise, thus improving segmentation performance. Furthermore, by masking the grouped global noise with local noise positions, the point frequency transformer can more effectively identify local noise, improving its local denoising capability.}


\subsection{Denoising PointNet}
The complex Point Frequency Transformer applied to high-resolution point clouds incurs a significant computational burden leading to long runtime.
To alleviate this problem, we propose a simple and efficient Denoising PointNet over the high resolution point cloud. It can be expressed as,
\vspace{-2mm}
\begin{equation}
\begin{aligned}
{f}_{i}&=\Lambda_{j=1}^{K} {\rm MLP} (G[f_i] \oplus P_{ij}) \\
&=\Lambda_{j=1}^{K}{\rm MLP}(f_{ij} \oplus P_{ij})\\
\end{aligned}
\vspace{-2mm}
\end{equation}
where $\Lambda_{j=1}^{K}$ is the maxpooling operation over $K$ neighborhood points. \textcolor{black}{ Incorporating the local position condition into the noisy features before applying MLP helps preserve spatial information and reduces the MLP's sensitivity to noise. This enhances the model's generalization and adaptability, while also reducing runtime compared to more complex local denoising aggregation methods.}


\begin{table}[t]
\centering
\scriptsize
\renewcommand\arraystretch{1.5}
\setlength{\tabcolsep}{0.2mm}{
\begin{tabular}{l|l|ccc|c|c|c|c|c}
\specialrule{1pt}{0pt}{0pt}
\cellcolor[gray]{0.9} 
&\cellcolor[gray]{0.9}  
&\multicolumn{3}{c|}{\cellcolor[gray]{0.9}{\textbf{S3DIS  6-fold }}} 
&\multicolumn{1}{c|}{\cellcolor[gray]{0.9}{\textbf{Area 5}}} 
&\multicolumn{1}{c|}{\cellcolor[gray]{0.9}{\textbf{Scan.}}}  
&\multicolumn{1}{c|}{\cellcolor[gray]{0.9}{\textbf{S.KI}}} 
&\multicolumn{1}{c|}{\cellcolor[gray]{0.9}{\textbf{SW.}}} 
&\multicolumn{1}{c}{\cellcolor[gray]{0.9}{\textbf{Shape.}}}\\\cline{3-10}

\multirow{-2}{*}{\cellcolor[gray]{0.9}\textbf{Year}} 
&\multirow{-2}{*}{\cellcolor[gray]{0.9}\textbf{Methods}}  
& \cellcolor[gray]{0.9}{\textit{ mIoU}}

& \cellcolor[gray]{0.9}{\textit{Para.}}   
& \cellcolor[gray]{0.9}{\textit{FLOPs}}  
& \cellcolor[gray]{0.9}{\textit{mIoU}} 

& \cellcolor[gray]{0.9}{\textit{mIoU}} 
& \cellcolor[gray]{0.9}{\textit{mIoU}} 
& \cellcolor[gray]{0.9}{\textit{mIoU}} 
& \cellcolor[gray]{0.9}{\textit{Ins. mIoU}}  \\\hline

\multicolumn{10}{l}{\textit{previous state-of-the-art point convolution based methods}} \\\hline
ICCV’19&KPConv\cite{thomas2019kpconv}  & 70.6       &15 &- &67.1 &69.2 &- &- &86.4    \\
CVPR’21 &PAConv~\cite{xu2021paconv}  & 69.3   &- &- &66.6  &- &- &- &86.1     \\
CVPR’24 &KPConvX~\cite{thomas2024kpconvx} &-     &13.5 &- &73.5 &76.3 &- &- &- \\\hline

\multicolumn{10}{l}{\textit{previous state-of-the-art MLP-based methods}} \\\hline

NIPS’22 &PointNeXt~\cite{qian2022pointnext}  &74.9     &41.6 &84.8 &71.1 &71.5 &- &50.4 &{87.2}    \\
CVPR’23 &PointVector~\cite{deng2023pointvector}  &{78.4}       &24.1 &58.5 &{72.3} &- &- &- & 86.9\\
CVPR`23 &PointMeta~\cite{lin2023meta}   &77.0     &19.7 &11.0 &{71.3}  &72.8 &- &- &87.1   \\
\hline

\multicolumn{10}{l}{\textit{previous state-of-the-art point transformer based methods}} \\\hline

ICCV’21 &PT~\cite{zhao2021point} & 73.5      &{7.8} &{5.6} &70.4  &70.4&-&58.6 &86.6  \\

CVPR'22 &ST~\cite{lai2022stratified}  &-  &- &-  &72.0 & 74.3 &- &- &86.6\\
NIPS’22 &PT V2~\cite{wu2022point}  &75.2    &12.8 &-  &71.6 &75.4 &70.3 &59.7&-  \\

NIPS’23 &ConDAF~\cite{duan2024condaformer}     &-  &-&-&\underline{73.5}      &76.0  &- &- &-\\

CVPR’24 &PT V3~\cite{wu2023point}   & {77.7}     &42.6 &- &73.1   &{77.5} &{70.8} &61.1&-  \\

TNNLS’24 & FPT~\cite{he2023full}  &76.8       &10.9 &8.3 &73.1 &75.6&69.6&\underline{61.8} &87.1\\\hline

\multicolumn{10}{l}{\textit{{diffusion based method}}} \\\hline
ICCV’23 &STPD~\cite{wu2023sketch} &-   &-   &- &-  &- &- &- & 86.7 \\
CASSP’24 &Liu~et~al~\cite{liu20243d}
&\underline{80.8}     &-  &- &- &- &- &- &\textbf{89.3}
\\
ArXiv’24 &CDSegNet~\cite{qu2024end}
&-   &-  &-  &-  &\underline{77.9} &- &- &-
\\
CVPR’24 &PointDif~\cite{zheng2024point}
&-   &-  &-  &70.0  &- &\underline{71.3} &- &-
\\

&\textbf{{PointDiffuse}}   &\textbf{81.2}      &15.2 &7.3  &\textbf{74.2}  &\textbf{78.2} &\textbf{71.4} &\textbf{64.8} &\underline{88.4} \\\hline
\end{tabular}}
\vspace{-2mm}
\caption{Segmentation results on 5 datasets. 
We report mIoU for S3DIS (6-fold and Area 5), ScanNet, SemanticKITTI and SWAN datasets. For part segmentation on ShapeNet, we report Ins. mIoU. We also report Parameters (million) and FLOPs (giga) for S3DIS dataset. 
In each column, the best result is bolded and second best is underlined. \% signs are omitted for clarity. Scan. = ScanNet; S.KI = SemanticKITTI; SW. = SWAN; Shape. = ShapeNet. All methods do not use the multi-dataset for joint training.}\label{table:five_dataset_main_results}
\vspace{-6mm}
\end{table}

\section{Experiments}
\subsection{Evaluation Setup}
\noindent\textbf{Datasets.} We conduct experiments on five popular datasets: S3DIS, ScanNetv2, SWAN, SemanticKITTI and ShapeNet. 
S3DIS \cite{armeni20163d} includes colored point clouds of 6 large-scale indoor scenes. Our evaluation employs Area 5 and 6-fold cross-validation on these 6 indoor scenes.
ScanNet~\cite{dai2017scannet} contains colored point clouds of indoor scenes. It is split into 1,201 scenes for training and 312 scenes for validation. 
SemanticKITTI~\cite{behley2019semantickitti} is an outdoor dataset using sequences 0 to 10 (excluding 8) for training and sequence 8 for validation.
SWAN~\cite{ibrahim2023sat3d} is a more recent outdoor dataset, where sequences 0 to 23 are allocated for training and sequences 24 to 31 are designated for testing. 
ShapeNet~\cite{yi2016scalable} is a 3D object part segmentation dataset, where 14,006 objects are used for training and 2,874 for testing. 

\noindent\textbf{Data Pre-processing.} \textcolor{black}{We follow data pre-processing strategies similar to previous methods\cite{zhao2021point,he2023full,wu2023point}}. For example, we subsample the input point cloud using the 4\textit{cm} grid for S3DIS and set the maximum number of subsampled points to 80,000. We adopt random scaling, random flipping, chromatic contrast, chromatic translation, and chromatic jitter to augment training data.  More details are  in \textit{Supplementary Material}. 

\noindent\textbf{Implementation Details.}
All experiments are conducted on the PyTorch platform and trained/tested on 4 NVIDIA 3090 GPUs. 
On the indoor S3DIS dataset, we pre-train the conditional network, gathering point labels from the training set of a single dataset due to limited computational resources without using multi-dataset joint training.
The denoising network is trained in an end-to-end manner using the AdamW optimizer with a batch size of 12. The initial learning rate is set to 0.5 and adjusted using MultiStepLR. Details are in \textit{Supplementary Material}.

\noindent \textbf{Evaluation Metrics.} 
We report three evaluation metrics including mIoU, parameters and FLOPs for S3DIS Area 5 experiment. On S3DIS 6-fold, ScanNet, SemanticKITTI and SWAN dataset, we report the mIoU metric and on ShapeNet, we report the Ins. mIoU metric.


\begin{figure}[t]
\centering
\includegraphics[width=3.3in ]{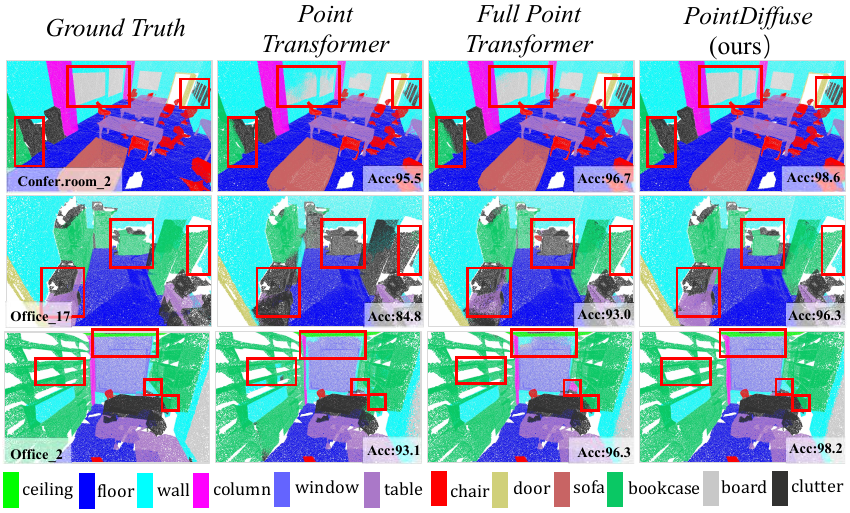}
\vspace{-4mm}
\caption{Semantic segmentation results on S3DIS Area-5. Red boxes indicate areas where our proposed PointDiffuse demonstrates noticeable improvements over previous methods, such as Point Transformer and Full Point Transformer.} 
\vspace{-4mm}
\label{fig:s3disVis}
\end{figure}

\subsection{Comparison with State-of-the-art Methods}
To verify the general point cloud segmentation performance, we compare PointDiffuse with recent state-of-the-art methods on five datasets without using multi-dataset joint training. Results are compiled in Table~\ref{table:five_dataset_main_results}.


\textcolor{black}{On the \textit{S3DIS} indoor dataset 6-fold validation, our method achieves a state-of-the-art performance of 81.2\% mIoU, surpassing the diffusion-based method~\cite{liu20243d} by 0.4\% mIoU. Our method outperforms Point Transformer~\cite{zhao2021point} (including V2~\cite{wu2022point} and V3~\cite{wu2023point}) and Full Point Transformer~\cite{he2023full} by large margins. On the S3DIS Area 5, our method also achieves the best performance of 74.2\% mIoU, outperforming the diffusion-based method (i.e., PointDif~\cite{zheng2024point}) by \textbf{4.2\% mIoU}. Furthermore, it exceeds the previous state-of-the-art MLP-based approach (i.e., PointVector~\cite{deng2023pointvector}) by \textbf{1.9\% mIoU}, while using only \textbf{63.1\%  parameters} and \textbf{12.5\% FLOPs}. Visualizations of our results on Area 5 are shown in Figure~\ref{fig:s3disVis}. }

On the \textit{ScanNet} indoor dataset, our method achieves the best result (i.e., 78.2\% mIoU) on the validation set, outperforming the diffusion-based method CDSegNet~\cite{qu2024end}.

On the \textit{SemanticKITTI} outdoor dataset, our method achieves 71.4\% mIoU on the validation set, outperforming transformer-based methods (e.g., Point Transformer V3 and Full Point Transformer) and diffusion-based method (i.e., PointDif~\cite{zheng2024point}).

On the  \textit{SWAN} outdoor dataset, we compare our results to the convolution-based method PointConv~\cite{wu2019pointconv}, MLP-based method PointNext~\cite{qian2022pointnext}, and transformer-based methods Point Transformer (including V2 and V3) and Full Point Transformer. These results are obtained by carefully training the models using the author-provided code and guidelines. Our PointDiffuse gets the best performance at 64.8\% mIoU, exceeding the nearest competitor, Point Transformer V3~\cite{wu2023point}, by \textbf{3\%}.

On the \textit{ShapeNet} object dataset, our method achieves a competitive performance of 88.4\% Ins. mIoU, compared to traditional deep learning models. Notably, our method outperforms the previous diffusion-based method STPD~\cite{wu2023sketch} by 1.7\%.

\textcolor{black}{Our results show that the proposed PointDiffuse outperforms all existing diffusion-based methods on the scene segmentation task but is only slightly worse than the method proposed by Liu et al.~\cite{liu20243d} in the part segmentation task. This is potentially because scene point clouds contain more noise compared to object point clouds. Since our method directly denoises labels, it is particularly effective in generating accurate labels for noisy scene point clouds.}

\begin{table}[t]

\centering
\arrayrulecolor{black}

\scriptsize
\renewcommand\arraystretch{1.3}
\setlength{\tabcolsep}{1.4mm}{
\begin{tabular}{c|cccc|cc}
\specialrule{1pt}{0pt}{0pt}
\rowcolor{gray!20} Case &{D-PN} &{NLE}/\textit{w} SC &{NLE}/\textit{w} PC &{PFT}    & S3DIS(\%)  &S.KITTI(\%)\\\hline
 I &\checkmark & & & &  71.6 &67.4 \\

 II &\checkmark & \checkmark & & &  73.2 &69.6 \\

 III &\checkmark &\checkmark &\checkmark & &73.7 &70.6 \\
 IV &\checkmark &\checkmark &\checkmark &\checkmark & \textbf{74.2} &\textbf{71.4} \\
 \hline

V &\checkmark &\checkmark & & \checkmark & {73.8} &{70.9} \\
VI &\checkmark & & \checkmark& \checkmark & {72.4} &{70.4} \\
\specialrule{1pt}{0pt}{0pt}
\end{tabular}
}
\vspace{-2mm}
\caption{Effect of various components. \textcolor{black}{D-PN: Denoising PointNet}. NLE = Noisy Label Embedding. /\textit{w} SC = with semantic condition. /\textit{w} PC = with position condition. PFT = Point Frequency Transformer. Metric: mIoU.}\label{tab:modules}
\vspace{-2mm}
\end{table}
\begin{table}[t]
\centering
\scriptsize
\renewcommand\arraystretch{1.4}
\setlength{\tabcolsep}{5.5mm}{
\begin{tabular}{l|ccc}
\specialrule{1pt}{0pt}{0pt}
\cellcolor[gray]{0.9}Backbone & \cellcolor[gray]{0.9}PT   &\cellcolor[gray]{0.9}PT V2 & \cellcolor[gray]{0.9}FPT   \\ \hline
step 5 &68.4 &71.9 &\cellcolor{orange!20}72.4  \\
step 10 &70.5 &74.3 &\cellcolor{orange!20}75.0 \\
step 15 &72.4 &76.3  &\cellcolor{orange!20}78.0  \\
\cellcolor{orange!20}step 20 &\cellcolor{orange!20}72.5 &\cellcolor{orange!20}76.4 &\cellcolor{orange!20}78.2 \\
step 25 &72.5 &76.4 &\cellcolor{orange!20}78.3\\ 
step 30 &72.5 &76.4 &\cellcolor{orange!20}78.4  \\\hline

\hline
\end{tabular}}
\vspace{-2mm}
\caption{Semantic segmentation on ScanNet. PT = Point Transformer. PT V2 = Point Transformer V2. FPT = Full Point Transformer. Metric: mIoU(\%).}\label{table:backbone}
\vspace{-4mm}
\end{table}

\subsection{Ablation Study}
We perform ablation experiments on the S3DIS and SemanticKITTI datasets to verify the effectiveness of the Denoising PointNet, Noisy Label Embedding, and Point Frequency Transformer in the PointDiffuse network. Additional ablation studies are in \textit{Supplementary Material}.

\vspace{0.5mm}
\noindent\textbf{Effect of Various Components.} \textcolor{black}{The effect of various components of} PointDiffuse are shown in Table~\ref{tab:modules}. \textbf{i)} Case I is the baseline simple PointDiffuse. Its denoising prediction network consists of the conditional network Full Point Transformer~\cite{he2023full} and simplified PointNet++. The \textcolor{black}{set abstraction module} in the PointNet++ is replaced with the proposed Denoising PointNet. \textbf{ii)} Cases II to IV systematically incorporate each of our proposed components, progressively enhancing the baseline results to reach 74.2\% and 71.4\% on the S3DIS Area 5 and SemanticKITTI validation set, respectively. \textbf{iii)} In Case V and VI, we remove the semantic and positions conditions from the noisy label embedding. Results indicate that the semantic condition is more crucial than the position condition.

\noindent \textbf{Effect of Conditional Network Backbone.} We study the impact of different backbones for the conditional network using Point Transformer~\cite{zhao2021point}, Point Transformer V2~\cite{wu2022point}, or Full Point Transformer~\cite{he2023full} in  Table~\ref{table:backbone}. PointDiffuse model with all three types of backbones achieve stable performance at step 20. However, using Full Point Transformer for the conditional network achieves the highest performance, with an accuracy of  78.4\% mIoU. Hence, we select  the Full Point Transformer as the backbone of conditional network, and adopt 20 diffusion steps to balance both accuracy and training efficiency for the ScanNet dataset.

\vspace{0.5mm}
\noindent \textbf{Effect of Loss.} We vary the hyper-parameter $\gamma$ in equation~\ref{eq:loss} and observe the performance of our network on S3DIS and SemanticKITTI. Table~\ref{tab:loss} shows that the best performance is achieved with 0.5 and 0.6, respectively. 

\begin{table}[t]
\centering
\arrayrulecolor{black}

\scriptsize
\renewcommand\arraystretch{1.3}
\setlength{\tabcolsep}{3.5mm}{
\begin{tabular}{c|cccccc}
\specialrule{1pt}{0pt}{0pt}
 \rowcolor{gray!20}$\gamma$ &0.3 &0.4 &0.5 &0.6  &0.7   \\\hline

 S3DIS(\%) &72.4 &73.0  &\textbf{74.2}    &72.4   &71.4 \\

 S.KITTI(\%) & 69.3 & 70.2 & 71.0 &\textbf{71.4} & 70.8 \\
\specialrule{1pt}{0pt}{0pt}
\end{tabular}
}
\vspace{-2mm}
\caption{Effect of hyper-parameter $\gamma$ on performance.}\label{tab:loss}
\vspace{-2mm}
\end{table}

\subsection{Further Analysis}

\noindent \textbf{Runtime Analysis.} \textcolor{black}{Table~\ref{table:dual_condition} shows results for our runtime analysis. The number and position of Denoising PointNet layers in the denoising network significantly influence training efficiency. The case I/II/III denotes the Denoising PointNet with the first one/two/three encoding layers and the last one/two/three decoding layers. Specifically, Case III, which places Denoising PointNet in the first three encoding layers and the last three decoding layers, achieves the shortest runtime—approximately 10 seconds per step.}

\begin{table}[t]
\centering
\scriptsize
\renewcommand\arraystretch{1.4}
\setlength{\tabcolsep}{2.5mm}{
\begin{tabular}{l|cc|cc|cc}
\specialrule{1pt}{0pt}{0pt}
\cellcolor[gray]{0.9} 
&\multicolumn{2}{c|}{\cellcolor[gray]{0.9}{\textbf{I }}} 
&\multicolumn{2}{c|}{\cellcolor[gray]{0.9}{\textbf{II }}} 
&\multicolumn{2}{c}{\cellcolor[gray]{0.9}{\textbf{III }}}\\\cline{2-7}

\multirow{-2}{*}{\cellcolor[gray]{0.9}\textbf{Case}} 
& \cellcolor[gray]{0.9}{\textit{mIoU}}
& \cellcolor[gray]{0.9}{\textit{Time/s}}  
& \cellcolor[gray]{0.9}{\textit{mIoU}}
& \cellcolor[gray]{0.9}{\textit{Time/s}} 
& \cellcolor[gray]{0.9}{\textit{mIoU}}
& \cellcolor[gray]{0.9}{\textit{Time/s}} \\\hline

step 5 &69.0 & 195      &68.8   & 109    &68.5  & 49\\
step 10 &71    & 390     &70.9  & 218    &70.6  &98\\
step 15 &73.8    & 585    &73.0  & 327    &72.6   &147\\
step 20 &74.6    &  780     &74.4    &436    &\cellcolor{orange!20}74.2                    &\cellcolor{orange!20}196\\
step 25 & 74.6   &  975    &74.4   & 545   &74.2  & 245\\ 
step 30 & 74.6   &  1170   &74.4   & 654  &74.3  &294 \\

\hline
\end{tabular}}
\vspace{-2mm}
\caption{\textcolor{black}{Segmentation accuracy and time (in sec) per scene on S3DIS Area 5 with different number of Denoising PointNet. s = second. }}\label{table:dual_condition}
\vspace{-4mm}
\end{table}

\noindent \textbf{Sampling Step Analysis.}  We study the impact of different sampling steps for point cloud segmentation on indoor, outdoor, and object datasets (see Figure \ref{fig:step}). \textcolor{black}{Depending on their complexity, different datasets require different numbers of sampling steps. We adopt 20, 30, and 15 steps for indoor, outdoor, and object datasets for optimal performance. We can observe that our network achieves the best performance with a few sampling steps on all datasets. Note that outdoor datasets require more steps than indoor datasets, and the object dataset requires the least steps. There are two potential reasons for this: \textbf{i)} Outdoor scenes contain a larger number of multi-scale objects and complex geometric structures, requiring more diffusion steps to gradually disentangle semantic-geometric relationships. \textbf{ii)} Noise in indoor datasets is closer to the standard Gaussian assumption, facilitating faster convergence.}

\begin{figure}[tbp]
\centering
\vspace{-2mm}
\includegraphics[width=3.2in ]{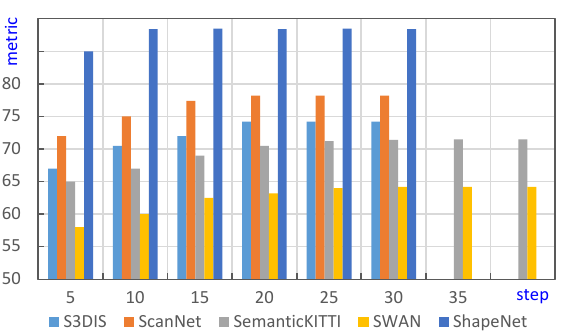}
\vspace{-2mm}
\caption{The effect of sampling step of PointDiffuse. We show the performance of metric (mIoU or Ins.mIoU) with increasing sampling steps.} 
\label{fig:step}
\vspace{-2mm}
\end{figure}

\begin{table}[t]
\centering
\scriptsize
\renewcommand\arraystretch{1.2}
\setlength{\tabcolsep}{2mm}{
\begin{tabular}{l|c|c|c|c|c}
\specialrule{1pt}{0pt}{0pt}
\cellcolor[gray]{0.9}Method & \cellcolor[gray]{0.9}mIoU   &\cellcolor[gray]{0.9}mAcc &\cellcolor[gray]{0.9} OA   &\cellcolor[gray]{0.9}Paras.(M) &\cellcolor[gray]{0.9}FLOPs(G) \\ \hline

PointNet~\cite{qi2017pointnet}        &47.6    &66.2    &78.5     &3.6 &35.5 \\
PointNet++~\cite{qi2017pointnet++} &54.5 &67.1 &81.0 &\textbf{1} & 7.2 \\

RandLA-Net~\cite{hu2020randla}        &70.0   &82.0   &88.0   &1.3  &5.8 \\
Point Transformer~\cite{zhao2021point}    &73.5     &81.9  &90.2   &7.8 &\textbf{5.6} \\
PointNext-XL~\cite{qian2022pointnext}       &74.9    &83.0  &90.3   &41.6 &84.8 \\
PointVector-XL~\cite{deng2023pointvector}       &78.4     &86.1  &91.9   &24.1  &58.5\\\hline
PointDiffuse/\textit{wt} STD/U   & \textbf{79.4}    &\textbf{87.3}   &\textbf{92.4}   &15.2 & 10.9 \\
\cellcolor{orange!20}PointDiffuse/\textit{w} STD/U  & \cellcolor{orange!20}\textbf{79.4}    &\cellcolor{orange!20}\textbf{87.3}   &\cellcolor{orange!20}\textbf{92.4}  &\cellcolor{orange!20}15.2 &\cellcolor{orange!20}7.3 \\

\hline
\end{tabular}}
\vspace{-2mm}
\caption{Semantic segmentation on S3DIS with 6-fold validation}
\vspace{-4mm}
\label{table:shared_transition_down_up}
\end{table}

\vspace{0.5mm}
\noindent \textbf{Analysis on Shared Transition Down/Up.} We conduct a comparative analysis between a noise prediction network employing shared Transition Down/Up and another utilizing Transition Down/Up while maintaining consistency in the remaining network configuration. Our results in Table~\ref{table:shared_transition_down_up} highlight the advantage of the Shared Transition Down/Up block as it achieves a similar performance to the one with Transition Down/Up block but with 33\% fewer FLOPs (7.3G vs 10.9G FLOPs). Note that the accuracy and number of parameters remain unchanged because the shared downsampling mechanism only shares neighborhood point indices to avoid redundant computations, without altering the learnable parameters of the network. 

\noindent \textbf{Analysis on Perturbation Robustness.} In Table~\ref{table:robustness}, we evaluate the robustness of our model to perturbations during testing. Following~\cite{xu2021paconv}, we apply perturbations such as permutation, rotation, shifting, scaling, and jitter to the point clouds. \textbf{i)} Our method achieves the best performance of 73.46\% mIoU without traditional data augmentation, demonstrating that the gains come from the diffusion model itself rather than the augmentation process. \textbf{ii)} our method is extremely robust to various perturbations. This further indicates that the diffusion model's ability to improve robustness comes from its denoising property. By removing Gaussian noise (which represents random perturbations), the diffusion model helps the model focus on the underlying data distribution rather than specific perturbation details. This denoising process trains the model to learn more robust features that are less sensitive to variations such as rotation, scaling, and jitter, thus improving performance under these perturbations.

\begin{table}[tbp]

\scriptsize
\centering
\renewcommand\arraystretch{1.5}
\setlength{\tabcolsep}{0.7mm}{
\begin{tabular}{l|c|c|ccc|cc|cc|c}
\specialrule{1pt}{3pt}{0pt}

\cellcolor[gray]{0.9}
&\cellcolor[gray]{0.9}
&\cellcolor[gray]{0.9}
& \multicolumn{3}{c|}{{\cellcolor[gray]{0.9}Rotation}}  
& \multicolumn{2}{c|}{{\cellcolor[gray]{0.9}Shifting}} 
& \multicolumn{2}{c|}{{\cellcolor[gray]{0.9}Scaling}}      
&\cellcolor[gray]{0.9}
\\ \cline{4-10}

\multirow{-2}{*}{{\cellcolor[gray]{0.9}Methods}} 
&\multirow{-2}{*}{{\cellcolor[gray]{0.9}None}}
&\multirow{-2}{*}{{\cellcolor[gray]{0.9}Perm.}}
&\cellcolor[gray]{0.9}$\pi$/2 
&\cellcolor[gray]{0.9}$\pi$ 
&\cellcolor[gray]{0.9}3$\pi$/2  
& \cellcolor[gray]{0.9}+~0.2   
& \cellcolor[gray]{0.9}-~0.2   
& \cellcolor[gray]{0.9}$\times$ 0.8  
&\cellcolor[gray]{0.9}$\times$ 1.2 
& \multirow{-2}{*}{{\cellcolor[gray]{0.9}Jitter}}   
\\ \hline
 PointNet++   &59.75  &59.71 &58.15 &57.18 &58.19 &22.33 &29.85  &56.24  &59.74 &59.04 \\
 {Minkowski}   &64.68  &64.56 &63.45 &63.83&63.36  &64.59  &64.96  &59.60  &61.93 &58.96   \\
PAConv &65.63 &65.64 &61.66&63.48&61.80  &55.81 &57.42 &64.20  &63.94 &65.12  \\
Point Tr.   &70.36 &70.45 &65.94&67.78&65.72  &70.44 &70.43 &65.73  &66.15 &59.67  \\
S.Tr.   &{71.96} &{72.02} &72.59&72.37&71.86  &{71.99} &{71.93} &{70.42}  &{71.21} &{72.02} \\
PointVe.   &{72.29} &{-} &72.27&72.30&72.32  &{72.29} &{72.29} &69.34 &69.26 &72.16 \\
{FPTr.} &{72.21}  &{72.23}  &- &- &- &{72.31} & {72.42} &{72.09}  &{71.48} &{72.31}   \\

\hline
\textbf{Ours} &\textbf{73.36}  &\textbf{73.35} &\textbf{73.76} & \textbf{72.94} & \textbf{73.26}  &\textbf{73.47} & \textbf{73.46} &\textbf{72.54}  &\textbf{73.10} &\textbf{73.38}  \\
\specialrule{1pt}{0pt}{0pt}

\end{tabular}
}
\vspace{-1mm}
\caption{Robustness study on S3DIS (mIoU\%) for random point permutations, $Z$ axis rotation ($\pi/2$, $\pi$, $3\pi/2$), shifting ($\pm~0.2$), scaling ($\times$0.8,$\times$1.2) and jitter in testing. Point Tr. = Point Transformer. S.Tr. = Stratified Transformer. PointVe. = PointVector. FPTr. = Full Point Transformer. }
\label{table:robustness}
\vspace{-4mm}
\end{table}

\section{Conclusion and Limitation}
\label{conclusion}
We proposed a novel dual conditional diffusion model-based network for point cloud semantic segmentation. To stabilize variance in the diffusion process, we introduced a Noisy Label Embedding mechanism, integrating both semantic and local position conditions. Furthermore, we designed a Point Frequency Transformer to effectively filter local geometric and global context noise. We developed a simple yet effective Denoising PointNet, which significantly reduces runtime. Extensive experiments across diverse datasets validate the robustness and efficacy of our approach.

\noindent \textcolor{black}{\noindent \textbf{Limitation:} Due to resource constraints, we have not yet been able to utilize multiple datasets for joint training of the conditional network. We believe that doing so could further enrich the learned representations and enhance the segmentation performance.}


\section{Acknowledgment}
This work was supported by National Key R\&D Program (2023YFB4704500), National Natural Science Foundation of China under Grant (U2013203, 62373140, U21A20487, 62103137), the Project of Science Fund for Distinguished Young Scholars of Hunan Province (2021JJ10024); Leading Talents in Science and Technology Innovation of Hunan Province (2023RC1040), the Project of Science Fund of Hunan Province (2022JJ30024); the Project of Talent Innovation and Sharing Alliance of Quanzhou City (2021C062L); the Key Research and Development Project of Science and Technology Plan of Hunan Province (2022GK2014). Professor Ajmal Mian is the recipient of an Australian Research Council Future Fellowship Award (project number FT210100268) funded by the Australian Government.
{
    \small
    \bibliographystyle{ieeenat_fullname}
    \bibliography{main}
}

\end{document}